\documentclass{article}

\usepackage{microtype}
\usepackage{graphicx}
\usepackage{subfigure}
\usepackage{booktabs} 

\usepackage{printlen}

\usepackage[utf8]{inputenc} 
\usepackage[T1]{fontenc}    
\usepackage{hyperref}       
\usepackage{url}            
\usepackage{booktabs}       
\usepackage{amsfonts}       
\usepackage{nicefrac}       
\usepackage{microtype}      
\usepackage{xcolor}
\usepackage{calc}

\usepackage{dsfont}    
\usepackage{todonotes}
\usepackage{amsmath,amssymb}
\usepackage{amsthm}
\usepackage{multirow}
\usepackage{graphicx}
\usepackage{listings}
\usepackage{microtype}
\usepackage{graphicx}
\usepackage{wrapfig}
\usepackage{booktabs} 
\usepackage{siunitx}
\usepackage{hyperref}
\hypersetup{
    colorlinks=true,
    linkcolor=black,
    filecolor=black,      
    urlcolor=blue,
    citecolor=black
}

\theoremstyle{definition}
\newtheorem{definition}{Definition}[section]

\newcommand{\mean}[2]{\mathbb{E}_{#2} \! \left[ #1 \right]}
\newcommand{\vect}[1]{\boldsymbol{#1}}
\newcommand\norm[2]{{\left\lVert#1\right\rVert}_{#2}}

\usepackage{hyperref}

\usepackage[accepted]{icml2024}

\usepackage{amsmath}
\usepackage{amssymb}
\usepackage{mathtools}
\usepackage{amsthm}

\usepackage[capitalize,noabbrev]{cleveref}

\theoremstyle{plain}

\theoremstyle{definition}

\theoremstyle{remark}

\icmltitlerunning{Stationarity without mean reversion in improper Gaussian processes}

\begin{document}

\twocolumn[
\icmltitle{Stationarity without mean reversion in improper Gaussian processes}



\icmlsetsymbol{equal}{*}

\begin{icmlauthorlist}
\icmlauthor{Luca Ambrogioni}{yyy}
\end{icmlauthorlist}

\icmlaffiliation{yyy}{Donders Institute for Brain, Cognition and Behaviour, Radboud University}

\icmlcorrespondingauthor{Luca Ambrogioni}{luca.ambrogioni@donders.ru.nl}

\icmlkeywords{Improper Gaussian Process, Mean Reversion, Non-Positive Kernel}

\vskip 0.3in
]



\printAffiliationsAndNotice{}  

\begin{abstract}
The behavior of a GP regression depends on the choice of covariance function. Stationary covariance functions are preferred in machine learning applications. However, (non-periodic) stationary covariance functions are always mean reverting and can therefore exhibit pathological behavior when applied to data that does not relax to a fixed global mean value. In this paper we show that it is possible to use improper GP priors with infinite variance to define processes that are stationary but not mean reverting. To this aim, we use of non-positive kernels that can only be defined in this limit regime. The resulting posterior distributions can be computed analytically and it involves a simple correction of the usual formulas. The main contribution of the paper is the introduction of a large family of smooth non-reverting covariance functions that closely resemble the kernels commonly used in the GP literature (e.g. squared exponential and Matérn class). By analyzing both synthetic and real data, we demonstrate that these non-positive kernels solve some known pathologies of mean reverting GP regression while retaining most of the favorable properties of ordinary smooth stationary kernels.
\end{abstract}

\section{Introduction}
Gaussian processes (GPs) are infinite-dimensional generalizations of Gaussian distributions that have gained widespread adoption in machine learning \citep{williams2006gaussian}, spatial statistics \citep{stein1999interpolation, gelfand2016spatial} and statistical signal processing \citep{tobar2015learning, lazaro2010sparse, ambrogioni2019complex, ambrogioni2018integral, tobar2023gaussian}. In machine learning, GPs offer a mathematically tractable way to study neural networks at the limit of infinitely many hidden units \citep{neal1996priors} and provide robust non-parametric models for low-dimensional regression and classification problems \citep{williams2006gaussian}. In statistical signal processing, GPs are used to specify structured prior distributions on the underlying signals, which can be used to specify dynamic properties such as temporal smoothness and oscillatory frequency \citep{tobar2015learning}. The properties of a GP depend on its covariance function, also known as a kernel. Depending on its covariance function, samples from a GP can exhibit a wide range of behaviors, such as polynomial growth, quasi-periodic oscillations, fractality or smoothness. In machine learning, the most commonly used covariance functions are stationary and isotropic. As we shall see in the paper, stationary covariance functions usually induce mean-reversion, meaning that perturbations tend to 'relax' towards a fixed average value. In this paper, we show that it is possible to use kernels that are both stationary and not mean-reverting as long as we use improper GP priors. Roughly speaking, an improper GP has infinite variance everywhere and its distribution of values at any point is given by an (improper) flat distribution. We introduce the use of a large family of \emph{non-positive kernels} that do not correspond to proper covariance functions.

\section{Preliminaries}
The use of improper priors is well-established in Bayesian statistics and machine learning. Loosely speaking, an improper prior is a non-normalizable density that nevertheless provides a proper normalizable posterior. Improper priors are often used as uninformative prior models. For example, consider a univariate Gaussian inference with prior $\mathcal{N}(x; 0, \nu^2)$ and likelihood $\mathcal{N}(y; x, \sigma^2)$. Given a data-point $y$, the resulting posterior distribution over $x$ is 
\begin{equation}
    p_{\nu}(x \mid y) = \mathcal{N}\! \left( x; \frac{\sigma^{-2}}{\nu^{-2} + \sigma^{-2}} y, (\nu^{-2} + \sigma^{-2})^{-1}\right)~.
\end{equation}
In this case, the posterior provides a biased estimator of the mean since the data $y$ is "shrunk" towards the origin (i.e. the prior mean). 
In order to remove the bias, we can take the limit $\nu^2 \rightarrow \infty$, which corresponds to an uninformative improper prior with infinite variance. However, while the prior does not exist as a proper probability distribution, the posterior has a well-defined (normalized) limit 
\begin{equation}
    p_{\infty}(x \mid y) = \mathcal{N}\! \left( x; y, \sigma^2\right)~,
\end{equation}
which provides an unbiased estimator of the mean.

\subsection{Gaussian process regression}
Consider a dataset comprised of $N$ pairs $\{(\vect{x}_1, y_1), \dots, (\vect{x}_k, y_k), \dots (\vect{x}_N, y_N)\}$, where $\vect{x}_k$ is a $D$-dimensional vector of predictors and $y_k$ is a scalar dependent variable. In a non-parametric regression problem, we assume the output variable $y_k$ to be sampled from an underlying unknown function $f: \mathds{R}^D \rightarrow \mathds{R}$ and corrupted with Gaussian white noise $\epsilon_k$ with variance $\sigma^2$:
\begin{equation}
    y_k = f(\vect{x}_k) + \epsilon_k~.
\end{equation}
The resulting non-parametric regression problem can be solved in closed-form if we assign a GP prior to the functional space:
\begin{equation}
    f(\vect{x}) \sim \mathcal{GP}\left( m(\vect{x}), k(\vect{x}', \vect{x}'')\right)~,
\end{equation}
where the prior process is defined by the mean function $m(\vect{x})$ and by the covariance function $k(\vect{x}', \vect{x}'')$. For the prior to be well-posed, the covariance function needs to be positive definite, meaning that all possible $M$-dimensional matrices obtained from it by evaluating it on a list of predictors are symmetric and positive-definite. In fact, if we restrict our attention on the set of predictors $\{\vect{x}_1, \dots, \vect{x}_N \}$, this functional prior reduces to a conventional multivariate Gaussian:
\begin{equation}
    \vect{f} \sim \mathcal{N}\left( \vect{m}, K\right)~,
\end{equation}
with $f_k = f(\vect{x}_k)$, $m_k = m(\vect{x}_k)$ and $K_{jk} = k(\vect{x}_j, \vect{x}_k)$. In the following, we will assume the mean function to be identically equal to zero. All results can be easily adapted to the case of a non-zero mean function. Given the dataset and an arbitrary set of $K$ queries $\{ \vect{x}_1^*, \dots, \vect{x}_K^*\}$, we can write down the posterior probability of the values of the function at the query points:
\begin{align}
    \vect{f}^* &\mid \{(\vect{x}_1, y_1), \dots, (\vect{x}_N, y_N)\} \\ \nonumber
    &\sim \mathcal{N}\left( \vect{m}_{\text{post}},  K_{\text{post}} \right)~,
\end{align}
with
\begin{equation}
    \vect{m}_{\text{post}} = K^{*} (K + \sigma^2 I)^{-1} \vect{y} 
\end{equation}
and
\begin{equation}
K_{\text{post}} = K^{**} - K^{*} (K + \sigma^2 I)^{-1} {K^*}^T~.
\end{equation}
In these expressions, the matrix $K^{**}_{jk} = k(\vect{x}^{*}_j,\vect{x}^{*}_k)$ is the (auto-)covariance matrix for the query set and the matrix $K^{*}_{jk} = k(\vect{x}^{*}_j,\vect{x}_k)$ is the cross-covariance matrix between the training set and the query set. 

\begin{figure}[h]\label{fig: symmetrized process}
\centering
\includegraphics[width=8cm]{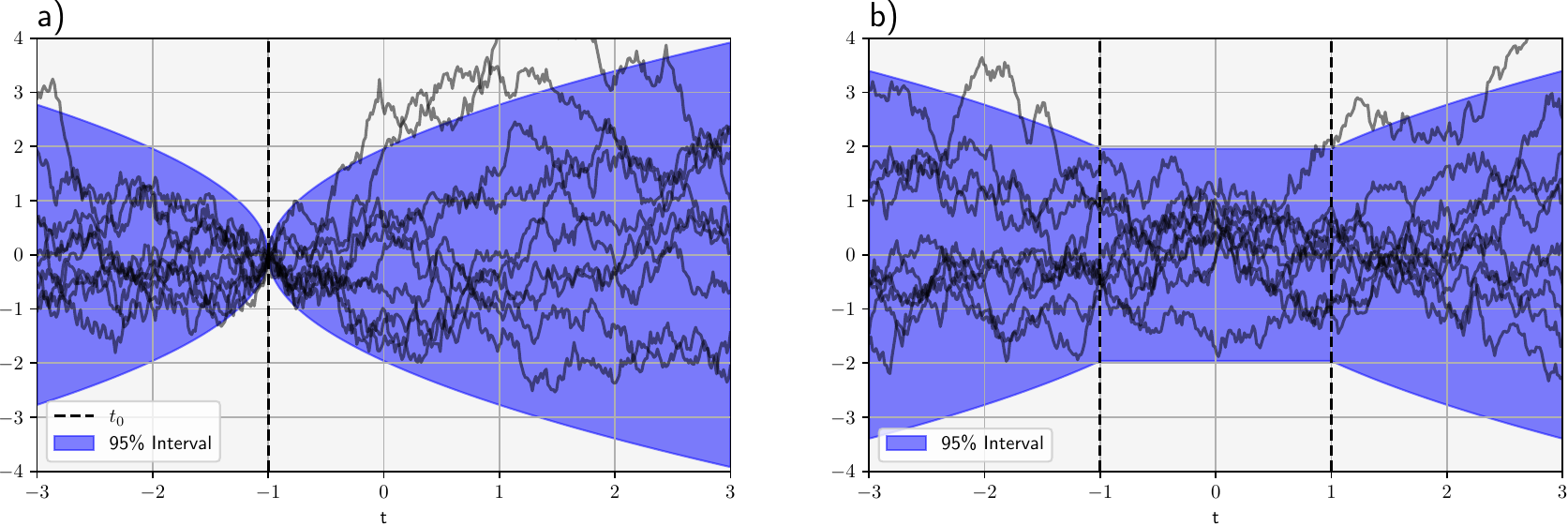}
\caption{Samples and $95\%$ probability intervals of a) bilateral Brownian motion starting at $-1$ and b) superposition of two bilateral Brownian motions starting at $-1$ and $1$ respectively. Note the 'stationary' interval with constant variance in between the two points.}
\end{figure}

\section{Stationarity and mean-reversion}
In the following sections we will restrict our attention to $1$-dimensional signals and we will denote the dependent variable as $t$. We will generalize most of the results to higher dimensional spaces in Sec.~\ref{sec: isotropic kernels}.

A covariance function $k: \mathds{R} \times \mathds{R} \rightarrow \mathds{R}$ is said to be stationary if it solely depends on the difference between the two input values:
\begin{equation}
    k(t,t') = \hat{k}(\tau)
\end{equation}
where $\tau = t' - t$ and $\hat{k}: \mathds{R} \rightarrow \mathds{R}$. Stationary covariance functions are extremely common in the machine learning literature since they do not require \emph{ad hoc} assumptions regarding the location-dependent statistical properties of the data. A well-known example of a stationary covariance function is the (misleadingly named) Squared Exponential:
\begin{equation}
    \hat{k}_{\text{se}}(\tau) = e^{- \frac{\tau^2}{2 l^2}}
\end{equation}
where $l$ is a length-scale parameter that regulates how fast the correlation between two points declines as a function of their distance. We say that a stationary covariance function is regular if $\hat{k}(\tau)$ is absolutely integrable, meaning that $\int_{0}^\infty |\hat{k}(\tau)| \text{d} \tau < \infty$. If the covariance function is regular, $\lim_{\tau \rightarrow \infty} \hat{k}(\tau) = 0$, since otherwise the integral would not converge. Stationary covariance functions are fully determined by their associated spectral density:
\begin{equation}\label{eq: spectral density}
    \tilde{k}(\omega) = \int_{-\infty}^{+\infty} \hat{k}(\tau) e^{-i \omega \tau} \text{d} \tau~,
\end{equation}
which is a real-valued function since the kernel is symmetric. The constraint of positive-definiteness implies that the spectral density is positive-valued. 

A covariance function is said to be mean-reverting if, for any $t$, the resulting stochastic process respects the following property
\begin{equation}
    \lim_{\tau \rightarrow \infty} \mean{f(t) \mid f(t - \tau)} {} = \mean{f(t)}{}~,
\end{equation}
In other words, in a mean-reverting process, the conditional expectation always converges to the unconditional expectation far away from the conditioning set. From the point of view of machine learning, this implies that the information provided by each data-point is local, since distant data points have negligible impact on the posterior distribution of a query point. While this assumption may seem reasonable, it does violate a basic principle of prediction: in the absence of further information, the most likely value of a variable in the future is its value in the present. For example, if our goal is to predict the performance of a student during college, given the fact that she has very high grades in primary school, it is safer to assume that the performance will on average stay high instead of assuming that it will revert to the population average. The assumption of mean reversion is also violated by many well-known and high-performance machine learning techniques. For example, in $k$-nearest neighbors regression and classification, the nearest training point is used to make a non-trivial prediction regardless of its distance from the query point \citep{taunk2019brief}. Similarly, methods such as binary trees, random forests and adaptive boosting exhibit a similar non-mean-reverting behavior \citep{ sutton2005classification, fawagreh2014random}. Finally, parametric regression models such as linear models, polynomial regression models and neural networks all tend to extrapolate their prediction away from the training data in a non-mean-reverting manner \citep{maulud2020review}. Given this, it is perhaps surprising that almost all applications of GP regression and classification to machine learning use mean-reverting covariance functions such as the Squared Exponential and the Matérn class. As we discussed above, the most likely reason for this choice is that stationary covariance functions tend to be less arbitrary and less dependent on domain knowledge. However, it is possible to show that all regular stationary covariance functions are mean-reverting (see supplementary section \ref{sup: stationarity} for an informal proof).

\begin{figure}[h] 
\centering \label{fig: mean reversion}
\includegraphics[width=8cm]{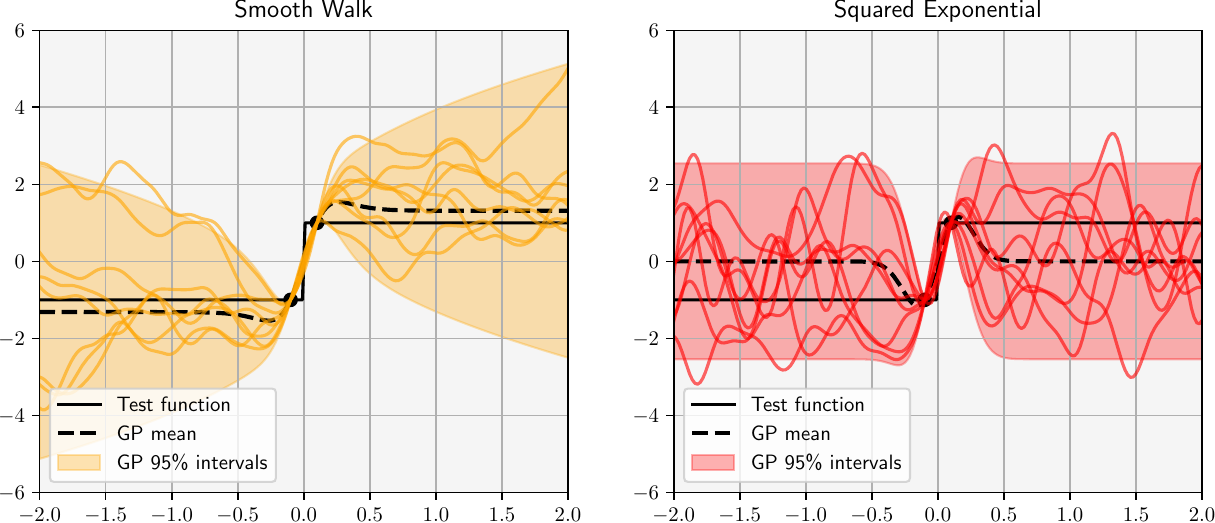}
\caption{Posterior expectation and $95\%$ intervals and samples of a GP conditioned on two data-points with A) improper Smooth Walk prior (top figure) and B) proper Squared Exponential prior. The length scale was $0.2$ for both models. The black line show the ground-truth signal $\text{sign}(t)$. }
\end{figure}

\begin{figure*}\label{fig: 1d test signal}
\centering
\includegraphics[width=14cm]{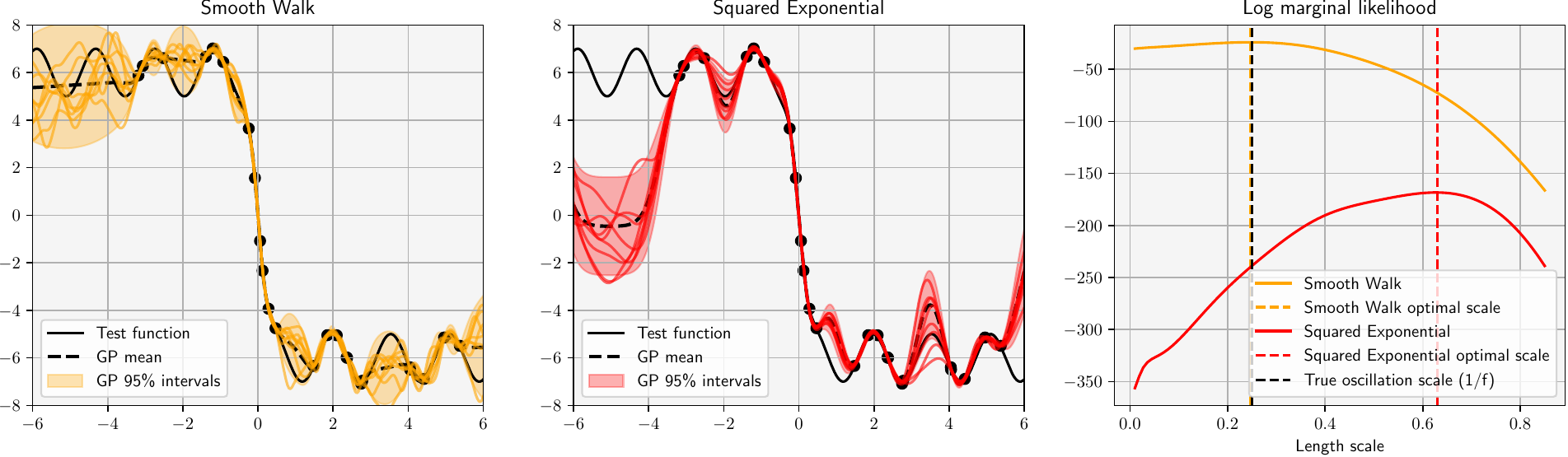}
\caption{Posterior expectation and $95\%$ interval of a GP regression with ground-truth signal $\sin(6 t) + 0.4 t - 5 \tanh(t)$ with noise level $\sigma = 0.05$. The length scale was selected by optimizing the marginal likelihood conditioned on one data point (see rightmost panel).}
\end{figure*}

\section{An intuitive introduction to stationary improper Gaussian processes}\label{sec: intuitive improper process}
Consider a (bilateral) Brownian motion process $z(t)$ that crosses the $y$-axis at $t_0$. This process is characterized by the following covariance function:
\begin{equation}
    k_{t_0}(t,t') = \frac{1}{2}\left( |t - t_0| + |t' - t_0| - |t' - t| \right)~.
\end{equation}
This process is not stationary since the covariance cannot be expressed as a function of the time difference $t' - t$. The lack of stationarity comes from the fact that the process is forced to cross the $y$-axis at $t_0$, which implies that the variance of the process is not translationally invariant. In fact, the variance of the process increases without bounds for $|t - t_0| \rightarrow \infty$ (see Fig.~1 a ). 

We can now consider the average of two bilateral Brownian processes with starting points equal to $c$ and $-c$ respectively, where $c$ is a free parameter. This leads to a new process characterized by the following covariance function:
\begin{equation}
    k_{\tau}(t,t') = \frac{1}{2}\left( k_{c}(t, t') + k_{-c}(t, t') \right)~.
\end{equation}
As shown in Fig.~2 b, the standard deviation of the process is constantly equal to $c$ for $|t| < c $ while it increases (decreases) for $|t| > c$. In particular, for $t, t' \in (-c, c)$, the covariance is given by $c - |t' - t| $. Therefore, the process is stationary when $t$ and $t'$ are restricted to be within this interval. We can now obtain a globally stationary limiting process by taking the limit $c \rightarrow \infty$, using a suggestive but improper notation, we can write the resulting covariance function as
\begin{equation}
    k_{\infty}(t,t') = \infty - |t' - t|~,
\end{equation}
where the infinity symbol reminds us that the variance of the process is infinite everywhere. Intuitively, this covariance describes a Brownian motion without a starting point. In the following section, we will formalize this intuitive reasoning in the context of improper GP regression. 

\section{Improper Gaussian process regression}
In this section, we will derive the formula for the posterior of an improper Gaussian process regression, where the variance of the prior process is infinite everywhere. We start by writing the covariance function of the process as follows:
\begin{equation} \label{eq: cov function}
    k(t, t') = s(t, t') + c
\end{equation}
where $c$ is an arbitrary number that will tend to infinity in the derivation. We denote the function $s(t, t')$ as the non-positive kernel of the process. In order to give rise to a well-posed improper GP prior, the function $s(t, t')$ needs to satisfy the following definition (see \cite{ong2004learning}):

\begin{definition}[conditionally positive definite kernel] \label{def: improper cov}
A $s(t, t'): \mathbb{R} \times \mathbb{R} \rightarrow \mathbb{R}$
 is a conditionally positive definite kernel when, for all $m \times m$ finite sub-matrices with $S_{ij} = s(t_i, t_j)$, we have that $S$ is symmetric and $\mathbf{u}^T S \mathbf{u} \geq 0$ when $\mathbf{u} \perp \mathbf{1}_m$, where $\mathbf{1}_m = (1, \dots, 1) \in \mathbb{R}^m$ is the constant one vector in the appropriate space. 
 \end{definition}

In the definition, $\vect{u} \perp \vect{1}_m$ means that $\sum_{j=1}^m u_j = 0$. Note that this property is less restrictive than the positive (semi-)definiteness. 

 From Def.~\ref{def: improper cov}, we can show that every $m$-dimensional covariance matrix obtained from Eq.~\ref{eq: cov function} is positive definite as far as the number $c$ is large enough. In fact, the covariance matrix can be decomposed as follows
\begin{equation}
    K = S + c \mathbf{1}_m\mathbf{1}_m^T
\end{equation}
where $\mathbf{1}_m$ is again a vector with constant elements all equal to $1$. Using the definition, we obtain
$$
\mathbf{u}^T K \mathbf{u} = \mathbf{u}^T S \mathbf{u} +  (\mathbf{u}^T \mathbf{1})^2 c~.
$$
Given a fixed vector $\mathbf{u}$, this quantity can be made positive for large enough values of $c$ since $\mathbf{u}^T S \mathbf{u}$ is finite and $(\mathbf{u}^T \mathbf{1})^2$ is always positive for non-zero vectors $\mathbf{u}$.

Let us now assume to have a $m$-dimensional vector $\mathbf{y}$ of observations of the dependent variable corresponding to the $m$ input points $\{t_1, \dots, t_m\}$. Assuming uncorrelated observation noise with variance $\sigma^2$ and that the value of $c$ has been chosen to make the covariance matrix positive-definite, the posterior expectation of the corresponding GP regression evaluated at the $n$ query points $\{t^*_1, \dots t^*_n\}$ is given by the formula
\begin{equation}\label{eq: gp posterior mean}
    \mu_{\text{post}} = (S^* + c \mathbf{1}_{n} \mathbf{1}_{m}^T) (\Sigma + c \mathbf{1}_m\mathbf{1}_m^T)^{-1} \textbf{y}~,
\end{equation}
where $S^*_{ij} = s(t^*_i, t_j)$ and $\Sigma = S + \sigma^2 I$. We can now obtain the expectation formula for the improper regression by taking the limit $c \rightarrow \infty$. Assuming $\Sigma$ to be invertible, this can be done by first applying the Sherman–Morrison formula to the matrix inverse:  
\begin{equation}\label{eq: Sherman–Morrison}
    (\Sigma + c \mathbf{1}_m\mathbf{1}_m^T)^{-1} = \Sigma^{-1} - c \frac{\Sigma^{-1} \mathbf{1}_m\mathbf{1}_m^T  \Sigma^{-1}}{1 + c \mathbf{1}_m^T \Sigma^{-1} \mathbf{1}_m}~.
\end{equation}
We can now plug Eq.~\ref{eq: Sherman–Morrison} in Eq.~\ref{eq: gp posterior mean} and take the limit, leading to the formula:
\begin{equation}\label{eq: improper posterior mean}
    \mu_{\text{post}} = S^* \Sigma^{-1} \mathbf{y} - S^* \frac{\Sigma^{-1} \mathbf{1}_m\mathbf{1}_m^T  \Sigma^{-1}}{\mathbf{1}_m^T \Sigma^{-1} \mathbf{1}_m} \mathbf{y} + \frac{\mathbf{1}_m \mathbf{1}_m^T \Sigma^{-1}}{\mathbf{1}_m^T \Sigma^{-1} \mathbf{1}_m} \mathbf{y}
\end{equation}
where the first term is identical to the posterior expectation formula of a regular GP regression except for the fact that $\Sigma$ is not necessarily positive-definite. Similarly, by taking the limit we can obtain a finite expression for the posterior covariance. To simplify the expression of the formula, we denote the matrix obtained by evaluating the conditionally positive definite kernel at the query points as $S^{**}_{ij} = s(t_j^*, t_k^*)$ and define the uncorrected covariance matrix as $\tilde{K}_{\text{post}} = S^{**} - S^* \Sigma^{-1} {S^*}^T$. We can now write the correct posterior covariance as the uncorrected covariance plus a correction term:
\begin{equation}
    \label{eq: improper posterior covariance}
    K_{\text{post}} = \tilde{K}_{\text{post}} + F F^T /(\vect{1}_m^T \Sigma^{-1} \vect{1}_m)
\end{equation}
where $F = \vect{1}_n - K^* \Sigma^{-1} \vect{1}_m$.

\begin{figure*}\label{fig: 2d test signal}
\centering
\includegraphics[width=15cm]{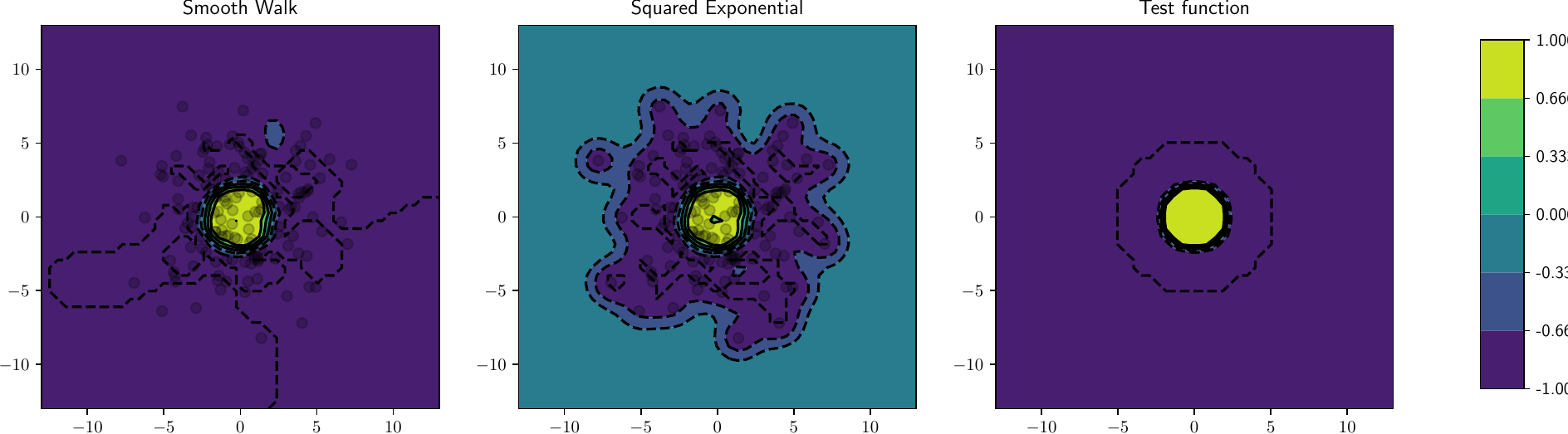}
\caption{Posterior expectation of a two-dimensional GP  regression with ground-truth signal (test function) $\tanh(-x^2-y^2 + 5)$. The posterior expectation with the Squared Exponential prior reverses to the prior mean while the bivariate Smooth Walk expectation stays close to the nearest data-point.}
\end{figure*}

\section{Stationary conditionally positive definite kernels}
Since any positive kernel automatically fulfils the conditionally positive definite kernel property, the formulas given in the previous section can be used together with ordinary covariance functions. In that case, like in the case of an ordinary improper Gaussian prior, the mean posterior expectation provides an unbiased (unregularized) estimate of the mean of the true signal. However, the biggest potential advantage of the approach comes from new conditionally positive definite kernels that are not positive definite and have properties that cannot be obtained with positive-definite kernels. In Section \ref{sec: intuitive improper process} we saw that, up to an infinite additive constant, the negative absolute difference $-|t' - t|$ can be seen as the (improper) kernel of a stationary Browning motion without starting points:
\begin{equation} \label{eq: neg dist kernel}
    s_{\text{bm}}(t, t') = \hat{s}_{\text{bm}}(\tau) = - |\tau|
\end{equation}
In words, the improper stationary kernel is simply the (negative) distance between the points. In Supp.~D, we prove that \ref{eq: neg dist kernel} respects definition \ref{def: improper cov}. The proof relies on the theory of tempered distributions, which is used to obtain the spectral density
\begin{equation} \label{eq: neg dist spectrum}
    \tilde{s}_{\text{bm}}(\omega) = \sqrt{\frac{2}{\pi}} \cdot \frac{1}{\omega^2}~,
\end{equation}
which can be interpreted as the distributional Fourier transform of the negative absolute value function \citep{strichartz2003guide} (see Supp.~A). However, it is important to keep in mind that $1/\omega^2$ is not a true density since it is not defined, and more importantly not integrable, at $\omega = 0$. This issue can be circumvented using a technique called Hadamard regularization (see Supp.~B), which can be used to assign a value to the divergent integral. However, this can lead to some rather counterintuitive behavior. For example, if we integrate Eq.~\ref{eq: neg dist spectrum} against a positive-valued test function on an interval including $\omega = 0$ we can obtain a negative result! This strange behavior is the reason why the kernel is not positive-definite and plays a central role in the proof. Samples from this kernel inherit the familiar properties of Brownian motions. As an example, since $\hat{s}(\tau)$ is continuous but not differentiable at the origin, the process is mean-squared continuous but not mean-squared differentiable. Importantly, like ordinary Brownian motions, the process is not mean-reverting.

More generally, we can show that the expectation does not revert to the mean for all kernels $k(\tau)$ that are asymptotic to $\tau$ fro $\tau \rightarrow \infty$. Consider a case with two data-points $y_{-}$ and $y_{+}$ corresponding to the coordinate values $-\alpha$ and $\alpha$. We evaluate the expected value at a query point $x^* = \alpha \pm \Delta$, where $\Delta$ is a finite number. We assume the improper kernel $k(\tau)$ to be asymptotically proportional to $-|\tau|$ for $\tau \rightarrow \infty$, which is true for all kernels introduced in our paper. To obtain a simple asymptotic formula, we start from Eq.~\ref{eq: improper posterior mean} and we take the limit $\alpha \rightarrow \infty$. In this limit, the two additive terms on the right hand side of Eq.~\ref{eq: improper posterior mean} individually converge to $-(y_{-} + y_{+})/2$ and $(y_{-} + y_{+})/2$ respectively, cancelling each other out at the limit. On the other hand, the term $S^* \Sigma^{-1} y$ converges to $y_{+}$ when $x^* = \alpha + \Delta$ while it converges to $y_{-}$ when $x^* = \alpha - \Delta$. Therefore, we have that

$$ \lim_{\alpha \rightarrow \infty} \mu_{\text{post}}(x \pm \Delta) = y_{\pm} $$

Importantly, this expression does not depend on the (finite) value of $\Delta$, which implies that the expectation is not mean-reverting as the posterior mean stays constantly equal to the value of the closest data-point.


\begin{table*} \label{tab: stock}
  \centering
    \begin{tabular}{c | c c c |  c c c }
    \toprule
    {} & SW (i) &  MW 1/2 (i) & GW (i) &  SE (p) &  M 1/2 (p) &  M 3/2 (p) \\
    \midrule
    NNL   & 10.34 &  5.89 & \textbf{3.90} &  76.02 &  74.23 &  74.78  \\
    SEM        & 2.09  & 0.79  & 0.07          &  21.35 &  21.58 &  21.19  \\
    \bottomrule
    \end{tabular}
    \caption{Quantitative results of the stock prices forecasting experiment. SW (i): Smooth Walk (improper), MW 1/2 (i): Matérn Walk 1/2, GW (i): Gaussian Walk, SE (p): Squared Exponential (proper), M 1/2: Matérn 1/2 (proper), M 3/2: Matérn 3/2 (proper).}
\end{table*}

\section{The Smooth Walk kernel}
In many applications it is beneficial to use smooth covariance functions capable of extrapolating linear and higher order trends in the data. In this section we will therefore introduce an infinitely smooth conditionally positive definite kernel that does not induce mean reversion. The idea is then to find a relaxation of the absolute value that is infinitely differentiable at the origin. The identity $|x| = x \text{sign}{x}$ suggests that we can do this by replacing the sign function with a hyperbolic tangent. This leads to the following kernel:
 \begin{equation}
     s_{\text{sw}}(t, t'; l) = -(t'-t) \tanh{\left( (t'-t)/ l \right)}
 \end{equation}
 where $l$ is a length scale parameter. We named it \emph{Smooth Walk} (improper) kernel since, as we shall see, it behaves like a smooth random walk. In Supp.~D, we prove that this is indeed an conditionally positive definite kernel for all positive values of $l$. The proof is analogous to the one discussed in the previous section and it relies on the fact that the distributional Fourier transform of $x \tanh{x}$ can be expressed using the density:
 \begin{equation}
     \tilde{s}_{\text{sw}}(\omega; l) = \frac{\pi^{3/2}}{2 \sqrt{2}} \frac{l^2 \coth(l~ \omega \pi /2) }{\sinh(l~ \omega \pi /2)}~,
 \end{equation}
which, like $1/\omega^2$, is not defined and not locally integrable at $\omega = 0$. In fact, we have that $\tilde{s}_{\text{sw}}(\omega; l) \sim \omega^{-2}/ (2 \sqrt{2})$ for $\lim \omega \rightarrow 0$. From this, it follows that $s_{\text{sw}}(\vect{x}, \vect{x}'; l)$ is not a positive-definite kernel, since the singularity in the density results in a (usually large) negative eigenvalue in the Gram matrices. Since $x \tanh{x}$ is infinitely differentiable at $t = 0$, the resulting process is infinitely mean-squared differentiable and exhibits very smooth samples. In this sense, the Smooth Walk conditionally positive definite kernel is similar the the Squared Exponential covariance function. However, like the Brownian motion kernel, the Smooth Walk kernel does not exhibit mean reversion, as can be seen in Fig.~2, which show the posterior expectation and intervals of a Smooth Walk and a Square Exponential process (with identical length scale) given two datapoints.

\section{Convolution kernels and the improper Matérn class}
In this section we will introduce a more general technique to construct conditionally positive definite kernels that replicate the behavior of existing stationary covariance functions. The main idea is to smooth out $-|t|$ by convolving it with a another (proper) kernel $\tilde{k}(\tau)$:
\begin{equation}
    \hat{s}_{\hat{k}}(\tau) = -\int_{-\infty}^{+\infty} |w| \hat{k}(\tau - w) \text{d} w
\end{equation}
Due to the convolution theorem, we know that the spectral density associated to this convoluted kernel is
\begin{equation} \label{eq: convolution spectrum}
    \tilde{s}_{\hat{k}}(\omega) = \sqrt{\frac{2}{\pi}} \frac{\hat{k}(\omega)}{\omega^2}~,
\end{equation}
where $\hat{k}(\omega)$ denotes the Fourier transform of $ \hat{k}(\tau)$, which is again singular at $0$ (at least if $\hat{k}(\omega) = 0$) and well-defined and non-negative valued everywhere else. This implies that Eq.~\ref{eq: convolution spectrum} gives us a valid (improper) kernel. 

We can now use this formula to construct conditionally positive definite kernels with different levels of smoothness. In general, if $\hat{k}(\tau)$ is $j - 1$-times differentiable at $0$, we have that $\hat{s}_{\hat{k}}(\tau)$ is $p$-times differentiable at the origin. This implies that the resulting process is $j$-times mean-squared differentiable. 

We can now introduce the following \emph{Matérn Walk} family of conditionally positive definite kernels:
\begin{equation} \label{eq: matern walk}
    \hat{s}_{j}(\tau) = -\int_{-\infty}^{+\infty} |w| \hat{\mathcal{M}}_{j - 1/2} ((\tau - w)/l) \text{d} w
\end{equation}
where $l$ is a length scale hyper-parameter and $\hat{\mathcal{M}}_{\nu} (\tau)$ is the Matérn class kernel, which is defined in terms of modified Bessel functions \citep{stein1999interpolation}. While the general expression involves special functions, this kernel greatly simplifies for $\nu = p + 1/2$ with $p$ being an integer. In this case, the kernel can be expressed as a product of an exponential and a polynomial of order $p$. Importantly, for this class of functions we can compute the convolution in closed-form. For example $\hat{m}_{1/2} (\tau/s) = e^{-|\tau|/s}$, which leads to the conditionally positive definite kernel:
\begin{equation} \label{eq: matern walk 1/2}
    \hat{s}_{1}(\tau) = -2 \left( |\tau| + l ~e^{-|\tau|/s} \right)~,
\end{equation}
which as expected is once differentiable at the origin. Another interesting special case is given by the limit $j \rightarrow \infty$. In this case, the Matérn converges to a Squared Exponential and the resulting conditionally positive definite kernel is
\begin{equation} \label{eq: gaussian walk}
    \hat{s}_{\infty}(\tau) = -\sqrt{\frac{2}{\pi}} e^{-\frac{\tau^2}{2 l^2}} - (\tau/l) ~ \text{erf}(\tau/(\sqrt{2} l))~,
\end{equation}
which is an improper version of the infinitely differentiable Squared Exponential kernel. As will refer to this kernel as \emph{Gaussian Walk} (GW) and to other members of the class as \emph{Matérn Walk} (MW).

\begin{table*} \label{tab: regression partial}
\centering \small
\begin{tabular}{llllll}
\toprule
0 &                                    3droad &                                    autompg &                                      bike &                             concreteslump &                                     energy \\
\midrule
GP SE (proper)                 &    $96.02 \pm 3.78$ &    $19.96 \pm 3.61$ &  $3.51 \pm 2.36$ &  $1643.39 \pm 216.09$ & $37.55 \pm 3.57$ \\
GP M 1/2 (proper)         &     $94.77 \pm 3.27$ &    $22.74 \pm 2.09$ &  $2.15 \pm 0.39$ &    $1726.44 \pm 316.93$ &     $34.57 \pm 3.66$ \\
GP M 3/2 (proper)         &    $92.92 \pm 2.45$ &   $31.57 \pm 11.16$ &    $5.03 \pm 1.51$ & $1683.64 \pm 82.48$ &    $40.83 \pm 2.34$ \\
GP M 5/2 (proper)         &    $88.27 \pm 2.356$ &     $17.96 \pm 1.72$ &   $4.06 \pm 1.81$ &   $1540.55 \pm 441.29$ &    $43.25 \pm 8.53$ \\
\midrule
Intrinsic Kriging ($\alpha = 1$) & $17.06 \pm 1.18$ & $2.23 \pm 0.23$ & $1.45 \pm 0.06$ &  $40.76 \pm 3.11$ & $2.75 \pm 
0.24$ \\
Intrinsic Kriging ($\alpha = 0.5$) & $21.18 \pm 0.74$ & $2.94 \pm 0.53$ & $\mathbf{1.40} \pm 0.03$ &  $39.04 \pm 2.87$ & $5.12 \pm 
0.84$ \\
\midrule
GP SW (improper)       &   $17.72 \pm 0.98$ &  $2.35 \pm 0.28$ &   $5.16 \pm 0.86$ &  $38.48 \pm 3.66$ & $3.00 \pm 0.17$ \\
GP MW 1/2 (improper) &  $14.37 \pm 0.69$ &  $\mathbf{1.79} \pm 0.04$ &  $4.85 \pm 0.3$ &  $35.28 \pm 1.61$ &   $\mathbf{2.41} \pm 0.14$ \\
GP GW (improper)     &  $\mathbf{4.01} \pm 0.16$ & $2.32 \pm 0.06$ &  $12.03 \pm 3.62$ &   $\mathbf{9.37} \pm 0.66$ &  $2.47 \pm 0.07$ \\
\bottomrule
\end{tabular}
\caption{Performance quantified as median negative log-likelihood on UCI regression datasets. To probe the low dimensional regime where stationary methods are the most appropriate, only the first $6$ predictive features of each dataset were used.
 Error bars are SE of the median (bootstrap).}
\end{table*}

\section{Improper isotropic kernels in high-dimensional spaces} 
\label{sec: isotropic kernels}
In this section, we will denote a $d$-dimensional vector as $\vect{x} \in \mathds{R}^d$. Under some regularity conditions, the conditionally positive definite kernels introduced in this paper can be generalized to higher dimensional spaces. A natural choice is to use an isotropic extension of the kernels. A kernel $k: \mathds{R}^D \times \mathds{R}^D \rightarrow \mathds{R}$ is said to be isotropic if the covariance of the process at two points can be expressed as a function of the Euclidean distance that separates them:
\begin{equation}
    k(\vect{x}',\vect{x}) = k_{\text{iso}}(\norm{\vect{x}' - \vect{x}}{2})~,
\end{equation}
where $\hat{k}_{\text{iso}} \mathds{R} \rightarrow \mathds{R}$. Under some conditions, it is possible to 'lift' a $1$-dimensional stationary kernel to a $d$-dimensional isotropic kernel. For example, we can define the isotropic version of the Brownian motion conditionally positive definite kernel as follows:
\begin{equation}
    s_{\text{bm}}^{(d)}(\vect{x}, \vect{x}') = \hat{s}_{\text{bm}}(\norm{\vect{x}' - \vect{x}}{2})~.
\end{equation}
Unfortunately, the resulting spectral density depends on the dimensionality and cannot easily be obtained from the univariate spectral density. This is potentially problematic since the resulting multivariate function is not guaranteed to be positive-definite or, in our case, a valid conditionally positive definite kernel. However, this can indeed be proven for improper stationary Brownian motion kernels. In fact, as proven in \citep{stein1971introduction}, the spectral density of $s_{\text{bm}}^{(d)}$ is given by the following distribution:
\begin{equation}
    \tilde{s}_{\text{bm}}^{(d)}(\vect{\omega}) = 2^{d+1} \frac{\Gamma((d+1)/2)}{\Gamma(-1/2)} {\norm{\omega}{2}^{-(d+1)}}~,
\end{equation}
which is again positive-valued up to a singularity at $\vect{\omega}$ that needs to be regularized. The Smooth Walk and Matérn Walk kernels can also be generalized into an isotropic kernel. Unfortunately, we were not able to obtain the spectral density analytically and we therefore cannot rigorously prove that this is indeed a valid conditionally positive definite kernel. However, we ran a comprehensive numerical analysis up to $d = 150$ and we found no violation of (improper) positive-definiteness (see supplementary section \ref{sup: numerical analysis}).

\section{Related work}
There is a substantial body of literature on how to avoid mean reversion in GP regression and Kriging by considering non-stationary models. For example, Majumdar and Gelfand proposed the use of non-stationary covariance matrices obtained using convolution operators \citep{majumdar2007multivariate}. Based on the theory of integrated white noise \citep{lindstrom1989fractional}, Zhang and collaborators constructed a non-mean-reverting non-stationary kernel obtained by integrating a smooth function of the Brownian motion kernel against white-noise from an arbitrary starting point \cite{zhang2015fractional}. The result was recently extended in \citep{zhang2016brownian} where the white-noise was replaced by a smooth GP, which leads to smoother covariance structures. As opposed to our current paper, these works adopt explicitly non-stationary kernels that assign different levels of prior variance to different locations. Similarly, several authors considered the use of explicit models of the trend that break the non-stationarity of the covariance matrix \cite{blight1975bayesian, journel1989we}. The approach was refined in \citep{raich2004orthogonal} where the effect of the covariance matrix is constructed to be orthogonal to the trend model in order to increase interpretability. 

The formulas for the corrected posterior given in Eq.\ref{eq: improper posterior mean} and Eq.\ref{eq: improper posterior covariance} are formally analogous to those used in the vague basis function approach as introduced in \citep{blight1975bayesian} and further developed in \citep{o1978curve}. In these works, a conventional GP kernel is used to fit the residuals of a least square polynomial regression. In this case, instead of taking the limit of the constant shift in the covariance matrix, the formulas are obtained by taking to limit of an improper prior for the coefficients of the basis functions. Nevertheless, in this work the kernel was taken to be positive-definite and only the prior on the basis functions was assumed to be improper. Our work is closely related to the theory of spline smoothing \citep{gu1993smoothing}. In fact, fitting a polynomial spline can be interpreted as a functional optimization problem in regularization theory \citep{kimeldorf1971some}, which can be reformulated as the MAP estimate of a GP regression with an improper prior on an infinite set of basis functions \citep{szeliski1987regularization}. The use of these improper priors induces a non-regularized functional subspace, which is defined as the null-space of the regularization operator. In our case, the non-regularized subspace is simply the space of constant functions. Differently from our conditionally positive definite kernels, the kernels implicitly defined by spline models are often non-stationary and therefore cannot be extended to high-dimensional spaces by isotropy. Furthermore, the functional operators used in spline theory are given by linear combination of a finite number of $p$-th derivatives and therefore cannot be used to define infinitely smooth kernels such as the Smooth Walk conditionally positive definite kernel introduced in this paper.

The theory of conditionally positive definite kernels have been studied in the spatial statistics, applied mathematics and signal processing literature \cite{matheron1973intrinsic, omre1987bayesian, handcock1993bayesian, ong2004learning, steinke2008kernels}. Specifically, the improper Gaussian processes framework used in this work can be fully formalized using the theory of \emph{intrinsic random function} (IRF) \citep{matheron1973intrinsic, cressie2015statistics}. From this perspective, our conditionally positive definite kernels can be seen as a special case of \emph{generalized covariance functions} \citep{matheron1973intrinsic}. The theory of intrinsic random functions has already found use in the kriging literature, based on the connection between intrinsic kriging and spline regression \cite{matheron1973intrinsic, vazquez2005intrinsic}. The intrinsic kriging approach assumes the differences $\vect{f}(\vect{x}) - \vect{f}(\vect{y})$ of an underlying stochastic function to be stationary, even when the function itself follows a non-stationary process (e.g. a Brownian motion). This leads to the use of non-positive definite generalized covariance functions defined as linear combinations of powers of $\norm{\vect{x} - \vect{y}}{2}$. The resulting posterior is equivalent to the posterior of our improper GP regression with a mixture of power-law conditionally positive definite kernels. However, these methods were not extended to more general smooth covariance structures such as the smooth walk and Matérn walk kernels introduced in this paper. 

Finally, similarly to our current paper, the limit kriging approach removes the effect of the prior mean using a limiting procedure \cite{joseph2006limit}. The paper analyzes the effect of this limit on stationary covariance functions and compares the results with intrinsic kriging. Note that removing the prior mean from an ordinary covariance function does not remove mean reversion, as the mean of the process will still revert to the mean of the data. Note that this is different from what is known as the flat limit, which refers to the divergence of the length scale of the kernel \citep{barthelme2023gaussian}.

\section{Experiments}
We start by showing the difference of behavior of GP regressions with proper and conditionally positive definite kernels on two appropriately chosen test functions in $1$d and $2$d (with isotropic kernels). Fig.~3 shows the result for the one-dimensional test function $\sin(6t) - 6 \tanh(4 t)$. The values of $t$ of the $25$ data points were sampled from a centered normal distribution with standard deviation equal to $2.5$. The noise standard deviation was $0.05$. As is clear from the figure, the GP regression with Squared Exponential covariance function fails to recover the correct length scale since, for low scales, the process erroneously reverts to the mean in between data points. Fig.~4 shows the result for the two-dimensional test function $\tanh(-x^2 -y^2 + 5)$. As we can see, the Smooth Walk kernel properly extrapolates the values to the whole plain outside the disk. On the other hand, the Squared Exponential extrapolation reverts to the global mean of $0$. \textbf{Probabilistic forecasting of stock prices.} We can now move on to the analysis of real world data. We consider a probabilistic forecasting problem where the aim is to predict the probability of future movements in a stock price given its past. Note that, due to market efficiency, it is extremely difficult to predict the expectation with any accuracy. The aim is then to have properly calibrated probabilistic intervals. $40$ timeseries of daily closing stock prices of SP-500 companies were selected at random. The data was log-transformed and then smoothed using a moving average filter with a time window of $35$ days. The first $100$ days were used as observation and the prediction was made over the following $300$ days. We performed the analysis using Squared Exponential (proper), Matérn class (proper), Smooth Walk, Matérn Walk and Gaussian Walk kernels. Since the actual marginal likelihood is divergent in improper models, the hyperparameters were optimized by minimizing the marginal likelihood conditional to one randomly selected observation. Table \ref{tab: stock} shows the mean and median predictive negative log-likelihood of all models. As clear from the table, the conditionally positive definite kernel greatly outperforms the Squared Exponential kernel. This is not surprising since stock prices are not thought to exhibit mean reversion. More details are given in Supp.~E. \textbf{Multivariate regression.}. The previous experiments have been chosen to highlight the potential shortcomings of mean reversion. In fact, stock prices offer a famous example of non-mean reverting timeseries. In this section, we compare the performance of GP repression with proper and conditionally positive definite kernels in a general multivariate regression benchmark, where it is unclear whether the assumption of mean reversion applies. Specifically, we test on regression problems from the UCI database \citep{uci1, uci2}. To probe the low dimensional regime, where stationary and isotropic methods are the most appropriate, only the first $6$ predictive features of each dataset were used. Furthermore, we considered the small data regime by sub-sampling (with replacement) $100$ datapoints from each training set. In all datasets, the training and test features were independently z-scored. We use isotropic extensions of the proper Matérn class and Squared Exponential kernels and the isotropic extensions of our improper Smooth Walk, Matérn Walk and Gaussian Walk kernels (see supplementary  \ref{sup: numerical analysis} for a numerical test of their validity).  We also included a comparison with two intrinsic kriging baseline models with power low kernels $k(\vect{x}, \vect{y}) = -\norm{\vect{x} - \vect{y}}{2}^p$ \cite{matheron1973intrinsic}, which were implemented using the spline fitting formulas given in \cite{williams2006gaussian} (Eq. 6.29, Chapter 6). These kriging models can also be seen as special cases of our approach, since power law kernels are conditionally positive definite kernels as we defined them in this paper. Like in the previous experiment, for all kernels, the length scale and noise level hyperparameters were optimized by maximizing the marginal log-likelihood conditional to one datapoint. Performance on the test set was quantified using the (univariate) negative log-likelihood. The results are given in table~\ref{tab: regression partial}. Conditionally positive definite kernels exhibit higher performance in all but one datasets. Table~\ref{tab: regression partial} shows the median negative log-probabilities calculated from $10$ bootstrap resamples of each training and test set. The error bars denote the standard error of the median computed using bootstrap resampling. The table shows that our improper method outperforms both the proper GP and the intrinsic Kriging baselines in all but one datasets. Table~\ref{tab: aggregated results} in Supp.\ref{sup: expt details} reports the aggregated results on a larger set of $36$ UCI regression datasets, which broadly confirms the results in table~\ref{tab: regression partial}. The supplementary material also includes the result of the aggregated results in the high-dimensional case, where we use the full number of features of each UCI regression dataset. As reported in Table.\ref{tab: aggregated results high d}, this analysis shows a larger performance gap between our conditionally positive definite kernels and the proper and intrinsic kriging baselines. In general, the Matérn Walk and Gaussian Walk kernels achieve higher performance than the Smooth Walk kernel. The power law kernels of the intrinsic kriging methods are often competitive, together with the less smooth Matérn 1/2 proper kernel. 

\section{Conclusions}
Our analyses show that improper GP regression with smooth conditionally positive definite kernel can be effectively used in many scenarios and it offers a competitive choice in tabular multivariate regression problems. Nevertheless, further theoretical work needs to be done to prove the kernel to be well-defined in higher dimensions, either by explicitly computing the spectral density or by proving that it respects the conditionally positive definite kernel property. Further work also needs to be done in order to construct and characterize larger families of kernels that satisfy the conditionally positive definite kernel property.

\section*{Impact Statement}
This paper presents work whose goal is to advance the field of Machine Learning. There are many potential societal consequences of our work, none which we feel must be specifically highlighted here.

\bibliographystyle{unsrtnat}
\bibliography{main.bib}

\appendix

\begin{table*} \label{tab: aggregated results}
\begin{tabular}{lrrrrrrrrr}
\toprule
{} &  SE (p) &  M 1/2 (p) &  M 3/2 (p) &  M 5/2 (p) &  SW (i) &  MW (1/2) (i) &  GW (i) &  IK 1 &  IK 1/2 \\
\midrule
\% Top score &        0.057 &                0.257 &                0.029 &                0.029 &                  0.057 &                        0.086 &                    \textbf{0.286} &       0.086 &         0.200 \\
Mean error  &        0.584 &                0.406 &                0.490 &                0.554 &                  0.234 &                        \textbf{0.169} &                    0.413 &       0.314 &         0.357 \\
SEM         &       49.092 &               49.157 &               49.066 &               49.064 &                  1.341 &                        1.251 &                    0.755 &       1.360 &         1.416 \\
\bottomrule
\end{tabular}
\caption{Aggregated results of on the UCI regression datasets using a maximum of six features (low dimensional regime). The error is quantified using the marginal negative log-likelihood on the test set.} 
\end{table*}

\begin{figure*}\label{fig: numerical}
\centering
\includegraphics[width=13cm]{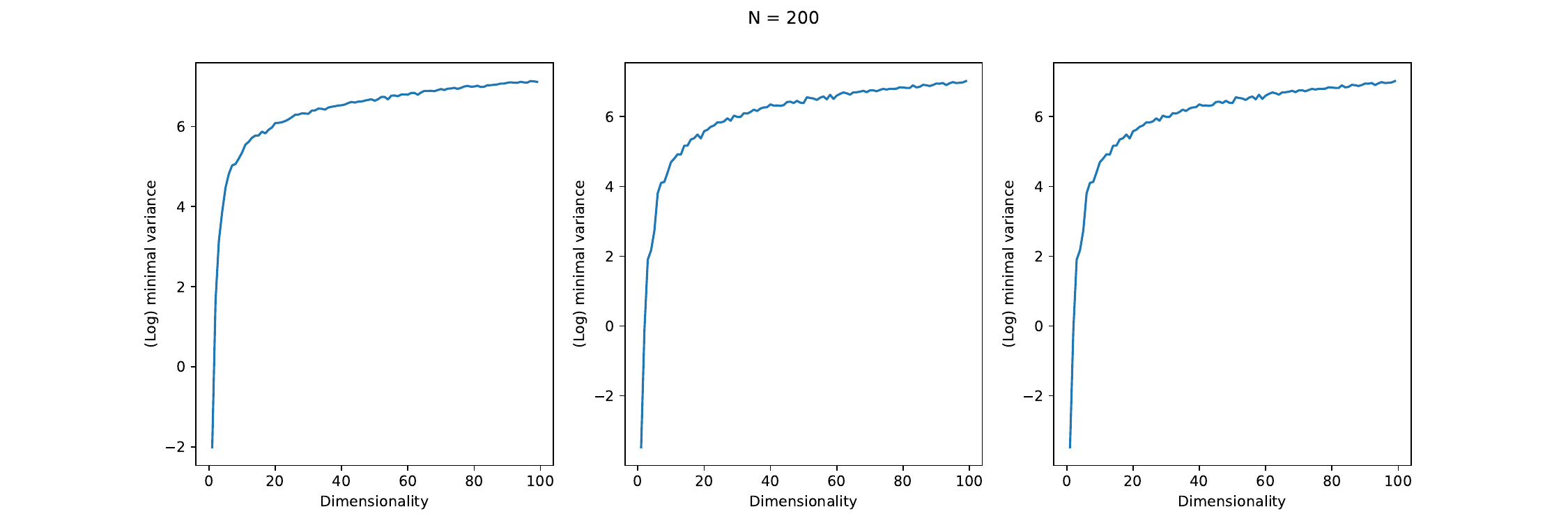}
\caption{Numerical test of (improper) positive-definiteness of smooth conditionally positive definite kernels.}
\end{figure*}

\section{Stationarity and mean reversion} \label{sup: stationarity}
It is easy to show that all regular stationary covariance functions are mean-reverting. Let us assume for simplicity that the mean function is identically equal to zero. In this case, the posterior mean function of a stationary GP regression (or classification) can always be written as follows
\begin{equation}
    \mu_{\text{post}}(t) = \sum_{j=1}^m w_j \hat{k}(t - t_n)~,
\end{equation}
where the $\vect{x}_n$ is a training point and $w_n$ is a real-valued weight. Since the covariance function is regular, we know that $\lim_{\tau \rightarrow \infty} \hat{k}(\tau) = 0$. This implies that $\mu_{\text{post}}(t)$ converges to the mean function (which we assumed to be identically equal to zero) as the distance between $t$ and the closest training point increases. This reasoning can be easily generalized to the case with a non-zero mean function. Note that this property does not necessarily hold for non-regular stationary covariance functions. In the machine learning literature, the most common examples of non-regular stationary covariance functions are the cosine and the periodic, which are both periodic with a spectral density defined as a mixture of delta distributions. In this case, the posterior expectation is itself periodic and it does not converge to any fixed value for $\tau \rightarrow \infty$, but instead it oscillates around the prior mean (or around a biased estimate of the data mean in the case of the periodic covariance functions).

\section{An informal introduction to the theory of tempered distributions} \label{sup: intro distributions}
Here, we will present an informal outline of the theory of tempered distributions. For a proper formal treatment, see \citep{gel2016generalized}.

Tempered distributions are a kind of 'generalized functions'. The main idea is to define these new functions as something that 'acts' on smooth test functions.

A the test functions $\phi: \mathds{R}^d \rightarrow \mathds{R}$ belonging to a space $\mathcal{S}[\mathds{R}^d]$ of rapidly decreasing functions defined on $\mathds{R}^d$. Loosely speaking, a function $\phi(\vect{x})$ is rapidly decreasing if it, together with all its partial derivatives of any order, tends to zero faster than any inverse power function for $\vect{x}$ for $\norm{\vect{x}}{2} \rightarrow \infty$. The rigorous definition of this \emph{Schwartz  space} is give, for example, in \citep{gel2016generalized}.

A tempered distribution $T$ is a functional $T: \mathcal{S} \rightarrow \mathds{R}$ that is both linear and continuous under an appropriate topology. Any locally integrable function (in the L$1$ sense) can be mapped into a distribution as follows:
 \begin{equation} \label{eq: integral representation}
     T_{g}[\phi] = \int_{\mathds{R}^d} g(\vect{x}) \phi(\vect{x}) \text{d} \vect{x}
 \end{equation}
 where the integral is interpreted in the Lebesgue sense. In this context, we refer to $g(x)$ as the density of the distribution. However, not all tempered distributions can be represented in this way. In this sense, tempered distributions generalize ordinary functions. A well-known example of a distribution that cannot be represented as an integral is the Dirac delta distribution, which simply evaluates the test function at a point:
  \begin{equation} \label{eq: dirac distribution}
     T_{\delta_{x'}}[\phi] = \phi(x')
 \end{equation}
The delta distribution is often written in the following sloppy but suggestive way
 \begin{equation}
     T_{\delta_{x'}}[\phi] = \int_{\mathds{R}^d} \delta(\vect{x} - \vect{x}') \phi(\vect{x}) \text{d} \vect{x}
 \end{equation}
 where the 'density' $\delta(\vect{x}$) is called Dirac delta function. Note however that this is just a notationally convenient shorthand for Eq.~\ref{eq: dirac distribution}. 

 The Fourier transform $\mathcal{F}[T]$ of a distribution $T$ is another tempered distribution defined by the following identity:
 \begin{equation}
     \mathcal{F}[T][\phi(\vect{x})] = T[\mathcal{F}[\phi(\vect{x})]], ~~~ \forall \phi \in \mathcal{S}~,
 \end{equation}
where
\begin{equation}
    \mathcal{F}[\phi(\vect{x})](\vect{y}) = \frac{1}{\sqrt{2 \pi}} \int_{\mathds{R}^d} \phi(\vect{x}) e^{-i \vect{x}^T \vect{y}} \text{d} \vect{x}
\end{equation}
is the ordinary (d-dimensional) Fourier transform of the test function $\phi(\vect{x})$. The basic idea behind this definition is to 'transfer' the definition of the Fourier transform from the test functions, where we can use the standard definition, to distributions, which are defined by their action on the test functions. Note that the definition agrees with the usual definition when the distribution comes from a regular function. In fact, for a distribution $T_g$ defined by a test function $g(\vect{x})$, we have
\begin{align}
    &T_g[\mathcal{F}[\phi(\vect{x})]] =  \\ \nonumber
    &\int_{\mathds{R}^d} g(\vect{x}) \left( \int_{\mathds{R}^d} \phi(\vect{x}) e^{-i \vect{x}^T \vect{y}} \text{d} \vect{x} \right) \text{d} \vect{y} \\ \nonumber
    &= \int_{\mathds{R}^d} \phi(\vect{x}) \left(\int_{\mathds{R}^d} g(\vect{x}) e^{-i \vect{x}^T \vect{y}}  \text{d} \vect{x} \right) \text{d} \vect{y} \\ \nonumber
    &= \mathcal{F}[T_g][\phi(\vect{x})] \\ \nonumber
\end{align}
where we could charge the order of integration due to the smoothness of both functions.

As we explained before, many tempered distributions cannot be represented as integrals weighted by a density function (Eq.~ \ref{eq: integral representation}). However, several important distributions that cannot be represented in this form can nevertheless be expressed as divergent integrals, together with a procedure for regularizing the infinite divergent term into a finite number. For example, consider the (one-dimensional) distribution:
\begin{equation}
        T_{1/x}[\phi(t)] = \mathcal{P} \int_{-\infty}^{\infty} \frac{\phi(x)}{x} \text{d} x
\end{equation}
where $\mathcal{P}$ denotes the Cauchy principal value of the divergent integral (see Supp.~\ref{sup: integral regularization}). This allows us to work with the singular density $1/x$, instead of dealing directly with the distribution. This allows us to easily generalize some familiar properties of the Fourier transform to this class of distributions. For our purposes, the most important is the convolution theorem. For example, assume that $f(x)$ is a regular smooth and absolutely integrable function and that $g(x)$ is a function such that $\mathcal{F}[g(x)](\omega) = 1/\omega + \tilde{h}(\omega)$, where $\tilde{h}(\omega)$ is again a smooth and absolutely integrable function. Then we can use the invoke the classical convolution theorem between functions:
\begin{equation}
    \mathcal{F}[f(x) \ast g(x)](\omega) = \tilde{f}(\omega) \left(1/\omega + \tilde{h}(\omega) \right)~.
\end{equation}
where $\tilde{f}(\omega)$ is the Fourier transform of $f(x)$. However, we need to keep in mind that the result will in general be another singular density and that we therefore need to properly regularize any integral involving $\mathcal{F}[g(x)] \mathcal{F}[g(x)]$ by taking the Cauchy principal value. 

The (partial) derivative of a tempered distribution is another tempered distribution, defined as follows
\begin{equation}
    \frac{\partial T}{\partial x_j}[\phi(\vect{x})] = - T\left[\frac{\partial \phi(\vect{x})}{\partial x_j} \right]~.
\end{equation}
This definition generalizes the usual integration-by-parts formula. In fact, consider $T_g$ to be the distribution associated to the non-singular density $g(x)$ defined on $\mathds{R}$, we have:
\begin{align}
   &\frac{\text{d} T}{\text{d} x}[\phi(\vect{x})] =  \int_{-\infty}^{\infty} \frac{\text{d} g(x)}{\text{d} x} \phi(x) \text{d} x \\ \nonumber
   &= \left[g(x) \phi(x) \right]_{-\infty}^\infty - \int_{-\infty}^{\infty} g(x) \frac{\text{d} \phi(x)}{\text{d} x} \text{d} x \\ \nonumber
   &= - \int_{-\infty}^{\infty} g(x) \frac{\text{d} \phi(x)}{\text{d} x} \text{d} x = - T_g\left[ \frac{\text{d} \phi(x)}{\text{d} x} \right]~,
\end{align}
where the boundary term vanishes since the test function is rapidly decreasing. This property, which is just a consequence of integration by parts, can then be generalized to define the derivative for distributions that cannot be written as integrals.

For example, the derivative of $f(x) = \frac{1}{2} \text{sign}(x)$ acts on a test function as follows
\begin{align}
&\frac{\text{d} T_{f}}{\text{d} x}[\phi(\vect{x})] = - T_g\left[ \frac{\text{d} \phi(x)}{\text{d} x} \right] \\ \nonumber
&= \frac{1}{2} \int_{-\infty}^\infty \text{sign}(x) \frac{\text{d} \phi(x)}{\text{d} x} \text{d} x \\ \nonumber
&= - \frac{1}{2} \left( \int_0^{+\infty}  \frac{\text{d} \phi(x)}{\text{d} x} \text{d} x - \int_{-\infty}^{0}  \frac{\text{d} \phi(x)}{\text{d} x} \text{d} x \right) \\ \nonumber
&= \phi(0)~,
\end{align}
where we used the fundamental theorem of calculus and the fact that test functions are rapidly decreasing. From the result, it is clear that the derivative of $T\frac{1}{2} \text{sign}(x)$ is just the Dirac delta distribution:
\begin{equation}
    \frac{\text{d} T_{\frac{1}{2} \text{sign}(x)}}{\text{d} x} = T_\delta~,
\end{equation}
which (using sloppy notation) can be expressed as:
\begin{equation}
    \frac{\text{d}}{\text{d} x} \text{sign}(x) = 2 \delta(x)~.
\end{equation}
Similarly, from the definition is clear that the derivative of the delta distribution extracts the derivative of the test function at $0$:
\begin{equation}
    \frac{\text{d} T_{\delta}}{\text{d} x}[\phi(x)] = -\phi'(0)~.
\end{equation}
which is often sloppily expressed by the density $\delta'(x)$. Note that this is just convenient notation, it is impossible to define this distribution $\delta'(x)$ by its values since, if the values were defined, they would be zero everywhere. In fact, for any interval $a, b$, we have
\begin{equation}
    \int_a^b \delta'(x) \text{d} x = \int_a^b \delta'(x) \cdot 1 \text{d} x = -\frac{\text{d} 1}{\text{d} x} = 0~.
\end{equation}

\begin{table*} \label{tab: aggregated results high d}
\begin{tabular}{lrrrrrrrrr}
\toprule
{} &  SE (p) &  M 1/2 (p) &  M 3/2 (p) &  M 5/2 (p) &  SW (i) &  MW (1/2) (i) &  GW (i) &  IK 1 &  IK 1/2 \\
\midrule
\% Top score &        0.029 &                0.200 &                0.029 &                0.086 &                  0.086 &                        0.229 &                    \textbf{0.257} &       0.029 &         0.057 \\
Mean error  &        0.397 &                0.430 &                0.354 &                0.358 &                  0.143 &                        \textbf{0.062} &                    0.183 &       0.501 &         0.499 \\
SEM         &       49.162 &               49.182 &               49.236 &               49.262 &                  1.336 &                        1.224 &                    1.194 &       1.252 &         1.335 \\
\bottomrule
\end{tabular}
\caption{Aggregated results of on the UCI regression datasets with the full number of features (high dimensionality). The error is quantified using the marginal negative log-likelihood on the test set.} 
\end{table*}

\section{Regularization of divergent integrals} \label{sup: integral regularization}
Here, we will discuss some techniques used to regularize divergent integrals. By 'regularize', we mean to extract a finite value from a divergent expression that normally returns an infinite value. 

Consider a function $f(x)/(x - x_0)$, where $f(x)$ is an absolutely integrable function. If $\lim_{x_0 \rightarrow 0} f(x) \neq 0$, improper integrals on an interval including zero do not converge. However, the Cauchy principal value of the integral is defined as follows
\begin{align}
    &\mathcal{P} \int_{a}^b \frac{f(x)}{x - x_0} \text{d} x \\ \nonumber
    &= \lim_{\epsilon \rightarrow 0} \left(\int_{a}^{x_0 - \epsilon} \frac{f(x)}{x - x_0} \text{d} x + \int_{x_0 + \epsilon}^b \frac{f(x)}{x - x_0} \text{d} x \right)~.
\end{align}
For example, it is easy to see that
\begin{equation}
    \mathcal{P} \int_{-1}^1 \frac{1}{x} \text{d} x = 0~,
\end{equation}
since the negative and positive divergences cancel each other out. As clear from the definition, the Cauchy principal value can only be used to regularize the integral of functions that diverge to a different sign of infinity around the singular point. Nevertheless, it is possible to regularize a larger family of divergent integrals using a technique known as Hadamard finite part. Consider a function $f(x)/(x - x_0)^2$, where $f(x)$ is an absolutely integrable function. The Hadamard finite part of the resulting integral can be obtained as follows
\begin{equation}
    \mathcal{H} \int_{a}^b \frac{f(x)}{(x - x_0)^2} \text{d} {x} = \left. \frac{d}{d a} \left( \mathcal{P} \int_{a}^b \frac{f(x)}{x - a} \text{d} x \right) \right|_{a = x_0}~.
\end{equation}
Note that the resulting finite part has some counterintuitive properties. For example, the result can be negative even if $f(x)/(x - x_0)^2$ is positive everywhere. For example, it is easy to show by direct calculation that
\begin{equation}
    \mathcal{H} \int_{-1}^{1} \frac{1}{x^2} \text{d} x = -2~.
\end{equation}
This strange result is central in understanding the properties of the conditionally positive definite kernels analyzed in this paper. In fact, these kernels 'fail' to be positive-definite due to this singular behavior of their associated spectral density. 

\section{Spectral densities of $-|x|$ and $-x \tanh(x)$}

We start by deriving the Fourier transform of $-|x|$. First of all, we write the function as
\begin{equation}
|x| = x ~\text{sign}(x)~,
\end{equation}
where $\text{sign}(x)$ is $1$ when $x > 0$, $-1$ when $x < 0$ and zero otherwise.

Since the derivative of the distribution $\text{sign}(x)$ is $2 \delta(t)$ (see Supp.~\ref{sup: intro distributions}) and $\mathcal{P} \int_{-\infty}^{+\infty} \text{sign}(x) \text{d} x = 0$, we have that
\begin{equation}
    - i \omega \mathcal{F}[\theta(x)](\omega) =  2 \mathcal{F}[\delta(x)](\omega) = \sqrt{2 / \pi}~,
\end{equation}
where $\theta(x)$ is the Heaviside step function, which gives
\begin{equation}
    \mathcal{F}[\text{sign}(x)](\omega) = \sqrt{\frac{2}{\pi}} \frac{i}{\omega}~.
\end{equation}
Now, since $\mathcal{F}[x f(x)](\omega) = \left. \frac{\text{d}}{\text{d} \xi} \mathcal{F}[f(x)](\xi)\right|_{\xi = \omega} $, we obtain:
\begin{equation}
    \mathcal{F}[-|x|] = - \mathcal{F}[x \text{sign}(x)] = \sqrt{\frac{2}{\pi}} \frac{i}{\omega^2}~.
\end{equation}
The Fourier transform of $-x \tanh(x)$ can be obtained in a similar way. The first step is to obtain the Fourier transform of $\tanh{x}$ as a distribution. This can be done by noticing that $\frac{\text{d} \tanh(x)}{\text{d} x} = \text{sech}^2(x)$, which is an absolutely integrable function whose Fourier transform can be obtained using usual techniques. In fact, it is possible to show that
\begin{equation}
    \frac{1}{\sqrt{2 \pi}} \int_{-\infty}^{+\infty} \text{sech}^2(x) e^{-i x \omega} \text{d} x = \sqrt{\frac{\pi}{2}} \omega ~\text{csch}(\pi \omega/ 2)~.
\end{equation}
From this, since $\mathcal{P} \int_{-\infty}^{+\infty} \tanh(x) \text{d} x = 0$, we have that
\begin{equation}
    \mathcal{F}[\tanh(x)](\omega) = i \frac{\mathcal{F}[\text{sech}^2(x)]}{\omega} = i \sqrt{\frac{\pi}{2}}~\text{csch}(\pi \omega/ 2)~.
\end{equation}
\begin{align}
    &\mathcal{F}[-|x|] = - \mathcal{F}[x \tanh(x)] = -i \sqrt{\frac{\pi}{2}} \left. \frac{\text{d}}{\text{d} \xi} ~\text{csch}(\pi \omega/ 2) \right|_{\xi = \omega} \\ \nonumber
    & = \frac{\pi^{3/2}}{2 \sqrt{2}} \text{coth}(\pi \omega/ 2) \text{csch}(\pi \omega/ 2).
\end{align}

\section{Proof that $-|\tau|$ and $-\tau \tanh(\tau)$ are conditionally positive definite kernels} \label{sec: proof}
In order to prove the statement, we need to introduce some concepts from the theory of tempered distribution \citep{gel2016generalized}. This theory allow to generalize the concept of functions and Fourier transforms, which will allow us to characterize the spectral density of the conditionally positive definite kernel. An informal review of the needed concepts is provided in Supp.~\ref{sup: intro distributions}.

We start by considering $\hat{s}_{\text{bm}}(\tau) = - |\tau|$. The symmetric property is trivially respected by all matrices. Therefore, we need to prove that for all $m \in \mathds{N}$, and all $m \times m$ matrices with $S_{jk} = -|t_k - t_j|$, we have that $\vect{u}^T S \vect{u} > 0$ if $\forall \vect{u}$ such that $\sum_j u_j = 0$.

For a given $m$-dimensional vector $\vect{u}$, we start by defining an associated smooth and rapidly decreasing function:
\begin{equation}
    v_{\sigma}^{\vect{u}}(t) = (2 \pi \sigma^2)^{-1/2} \sum_{j}^m u_j e^{-\frac{(t - t_j)^2}{2 \sigma^2}} 
\end{equation}
We can now write the finite matrix product as a limit of the following double integral:
\begin{equation}
    \vect{u}^T S \vect{u} = \lim_{\sigma \rightarrow 0} \int_{-\infty}^{+\infty} \int_{-\infty}^{+\infty} v_{\sigma}^{\vect{u}}(t) \hat{s}(t' - t) v_{\sigma}^{\vect{u}}(t') \text{d} t \text{d} t'
\end{equation}
Since the kernel is a function of $t' - t$, we can re-express the integral as a (functional) inner product:
\begin{equation}
    \int_{-\infty}^{\infty} v_{\sigma}^{\vect{u}}(t) ~\hat{s} \ast  v_{\sigma}^{\vect{u}} (t) \text{d} t~,
\end{equation}
where $\hat{s} \ast  v_{\sigma}^{\vect{u}} (t)$ denotes the convolution between the kernel and the function $v_{\sigma}^{\vect{u}}(t)$. We can now invoke Plancherel theorem and the convolution theorem (See Supp.~\ref{sup: intro distributions}) to re-write the inner product in the Fourier domain:
\begin{equation}\label{eq: spectral limit}
    \vect{u}^T S \vect{u} = \lim_{\sigma \rightarrow 0} \mathcal{H} \int_{-\infty}^{+\infty} |v_{\sigma}^{\vect{u}}(\omega)|^2 \tilde{s}(\omega) \text{d} \omega
\end{equation}
where $v_{\sigma}^{\vect{u}}(\omega)$ is the ordinary Fourier transform of $v_{\sigma}^{\vect{u}}(t)$ and $\mathcal{H}$ denotes the fact that the singular part of the divergent integral needs to be regularized by taking the Hadamard finite part. Note that the divergence of the ordinary integral comes from the fact that the Fourier transform of $-|\tau|$ can only be expressed as a tempered distribution. 

For $\hat{s}_{\text{bm}}(\tau)$, we have that the spectral density is proportional to $1/\omega^2$. On the other hand, it is easy to check that, if $\sum_j^m u_j = 0$, $|v_{\sigma}^{\vect{u}}(\omega)|^2 \sim \alpha \omega^2$for $\omega \rightarrow 0$, with $\alpha$ being a factor depending on $\sigma$. This implies that $|v_{\sigma}^{\vect{u}}(\omega)|^2$ has a finite limit at $\omega = 0$ and, consequently, that the integral in Eq.~\ref{eq: spectral limit} does not have divergences that require Hadamard regularization. This proves the statement, since an integral of a smooth positive-valued function is always non-negative-valued and a limit of a non-negative-valued series is always non-negative. Note that the spectral density can be negative when $\sum_j^m u_j \neq 0$ since the resulting regularization of the divergence at $\omega = 0$ can result in negative values (see Supp.~\ref{sup: integral regularization}). The result straightforwardly apply to $-\tau \tanh(\tau)$ as well since its spectral density is asymptotically equivalent to $1/\omega^2$ for $\omega \rightarrow 0$ and it is smooth everywhere else.

\section{Numerical analysis of conjectured isotropic kernels} \label{sup: numerical analysis}
In this paper, we conjecture that the Smooth Walk, Gaussian Walk and Matérn Walk kernels introduced in section \ref{sec: isotropic kernels} are valid improper covariance functions for any dimensionality, meaning that they respect Def.~\ref{def: improper cov}. Unfortunately, we were not able to prove this conjecture using analytic methods since we could not express the spectral density distribution associated to these isotropic kernels (for $d$ > 1) in closed-form. To corroborate our conjecture, we run a series of numerical experiments that could to falsify our hypothesis. Specifically, for any dimensionality from $d = 2$ to $d = 100$ we constructed a $200 \times 200$ covariance matrix by randomly sampling $200$ query points from independent uniform distributions in $(-1, 1)$. We then randomly sampled a $10000 \times 200$ matrix of random vectors sampled a standard normal distribution. These vectors were then numerically z-scored in order to fulfil the zero mean condition required by Def.~\ref{def: improper cov}.
Using these normalized vectors, we evaluated $10000$ inner products using the formula
\begin{equation}
    v = \vect{w}^T K \vect{w}~.
\end{equation}
In all the considered kernels, we did not find any violation of the positivity condition of the inner products up to double numerical tolerance. The minimum of the inner products as function of the dimensionality is visualized in Fig.~\ref{fig: numerical}. As you can see, the estimated minimum seems to increase monotonically as function of the dimensionality. We repeated the same experiments for matrix sizes of $10$, $20$ and $100$ without finding any violation in any of the kernels.

\section{Details of the time series experiment} \label{sup: stock details}
Time series of stock prices (daily closing time) were extracted using the yfinance (yahoo finance) python package. The stocks were selected from the list of S\&P 500 companies as it was on June 2023. We extracted time series from the perios danging from 2010-01-10 to 2021-03-03. The closing prices were log transformed and then filtered using a causal moving average of $35$ days. For each timeseries, we trained the model on the first $100$ time points and we evaluated the probabilistic prediction on the following $1000$ time points. This split allowed us to evaluate the probabilistic calibration of long term forecasting. The noise level and length scale of both proper and conditionally positive definite kernels was optimized by maximizing the marginal log-likelihood conditional on a random time point. Performance was quantified by evaluating the predictive marginal negative log likelihood of each model.  

\section{Details of the multivariate regression experiments} \label{sup: expt details}
GP regression were performed using the exact formulas. For scalability reason, if the $n$ in a dataset was larger than $100$ datapoints. We used a sub-training set sampled without replacement from the original dataset. This also allowed us to train on several re-samplings of the same dataset so as to evaluate error bars. Since we mostly cared about the low-dimensional regime, which is more appropriate for isotropic methods, we only used the first $6$ features of each dataset (assuming that they had more than $6$ features. However, we also included results analyzed using all the features of each datasets. All kernels were parameterized by a length scale and a noise std parameter, which were optimized by maximizing the log-marginal likelihood conditional to one, randomly chosen, datapoint. The intrinsic kriging baselines were implemented using the equivalent spline fitting formula, which is also a special case of our posterior formula. 

We tested the different kernels a a selection of UCI datasets suitable for regression. We used the following datasets: 3droad	(n = 434874, d = 3);  autompg (n=392, d=7); bike (n=17379, d=17); concreteslump	(n=103, d=7); energy (n=768; d=8); forest (n=517, d=12); houseelectric(n=2049280, d=11);  keggdirected (n=48827, d=20); kin40k(n=40000, d=8); parkinsons (n=5875, d=20); pol (n = 15000, d=26); pumadyn32nm (n=8192; d=32); slice (n=53500, d=385); solar (n=1066, d=10); stock (n=536; d=11); yacht (n=308, d=6); airfoil (n=1503, d=5); autos (n=159, d=25); breastcancer (n=194, d=33);
buzz (n=583250,	d=77); concrete (n=1030, d=8); elevators (n=16599, d=18); fertility (n=100, d=9); gas (n=2565, d=128); housing	(n=506, d=13); keggundirected	(n=63608, d=27); machine (n=209,d=7); pendulum (n=630, d=9);  protein (n=45730, d=9); servo (n=167, d=4); skillcraft (n=3338, d=19); sml	(n=4137, d=26); song (n=515345, d=90); tamielectric (n=45781, d=3); wine (n=1599, d=11).

The aggregated results in the low dimensional case are given in table~\ref{tab: aggregated results}, while the results in high dimension are given in table~\ref{tab: aggregated results high d}.



\end{document}